\documentclass[runningheads]{llncs}
\usepackage{graphicx}
\usepackage{comment}
\begin{document}
\title{The URW-KG: a Resource for Tackling the Underrepresentation of non-Western Writers}

\titlerunning{The URW-KG}
%
\author{Marco Antonio Stranisci\inst{1}\orcidID{0000-0001-9337-7250} \and
Giuseppe Spillo\inst{2}\orcidID{0000-0001-8345-4232} \and
Cataldo Musto\inst{2}\orcidID{0000-0001-6089-928X} \and Viviana Patti\inst{1}\orcidID{0000-0001-5991-370X} \and Rossana Damiano\inst{1}\orcidID{0000-0001-9866-2843}}
\authorrunning{M. A. Stranisci et al.}
%
\institute{University of Turin. Corso Svizzera 185, Turin, Italy \and
University of Bari Aldo Moro. Via Orabona 4, Bari, Italy
\email{\{marcoantonio.stranisci,viviana.patti,rossana.damiano\}@unito.it}\\
\email{\{giuseppe.spillo,cataldo.musto\}@uniba.it}}

%
\maketitle              
\begin{abstract}
Digital media have enabled the access to unprecedented literary knowledge. Authors, readers, and scholars are now able to discover and share an increasing amount of information about books and their authors. Notwithstanding, digital archives are still unbalanced: writers from non-Western countries are less represented, and such a condition leads to the perpetration of old forms of discrimination. In this paper, we present the Under-Represented Writers Knowledge Graph (URW-KG), a resource designed to explore and possibly amend this lack of representation by gathering and mapping information about works and authors from Wikidata and three other sources: Open Library, Goodreads, and Google Books. The experiments based on KG embeddings showed that the integrated information encoded in the graph allows scholars and users to be more easily exposed to non-Western literary works and authors with respect to Wikidata alone. This opens to the development of fairer and effective tools for author discovery and exploration.

\end{abstract}
\section{Introduction}

In the last few years, digital media have impacted on the fruition of literary works by unfolding new opportunities. New practices of reading were born through social platforms like Goodreads\footnote{\url{https://www.goodreads.com/}} \cite{nakamura2013words}, and the relationship between writers and their audience was reshaped, driving cultural and economic transformations \cite{shaver2020books}. 
The diffusion of large open sources of knowledge \cite{stroube2003literary,vrandevcic2014wikidata} has engaged literary scholars in new research issues and practices \cite{gil2016global} which rely on Semantic Web technologies \cite{tillett2005frbr,brown2014scaling}. 
Such a transformation has influenced also computer scientists, who have exploited literary works data stored in self-publishing platforms \cite{zhu2015aligning} and digital archives \cite{brooke2015gutentag} to train models for Natural Language Processing \cite{devlin2018bert}.

The impact of digital media on the literary ecosystem is not free from flaws, though. The ecosystem of digital archives of literature is vast, but fragmented, and not all resources acknowledge the Linked Data paradigm. 
For instance, there is no systematic mapping of writers' pages on Wikidata onto other sources such as OpenLibrary\footnote{\url{https://openlibrary.org}} and Worldcat\footnote{\url{https://www.worldcat.org}}. Since these knowledge bases have proven to be flawed by the lack of neutrality, such a limitation is even more critical as it hinders their comparative analysis.
For instance, as highlighted in some recent studies, Wikipedia include biases \cite{field2022controlled} as well as a lack of information about non-Western people \cite{adams2019counts}. Such underrepresentation reduces the possibility of discovering, identifying, and suggesting non-Western writers and their works both to the general public and to domain experts like scholars in Digital Humanities (DH).

In this paper, we address this issue through the creation of the Under-Represented Writers Knowledge Graph (URW-KG), a dataset of writers and their works targeted at assessing and  reducing  their potential lack of representation.
The URW-KG has been designed to support the following research objectives (ROs):
\begin{enumerate}
    \item Exploring the underrepresentation of non-Western writers in Wikidata by aligning it with external sources of knowledge; 
    \item Supporting the research in the humanities and the scholars in Digital Humanities with an unbiased digital archive of works and writers;
    \item  Fostering the exploration and discovery of potentially underrepresented writers through applications built on top of the KG.
    \end{enumerate}

In order to validate the intuitions that drove the design and the development of the KG, we used the triples encoded in the graph to feed several graph embedding models \cite{Cai:18}. The resulting vectors have been used to assess to what extent our KG allows discovering under-represented authors.


The paper is structured as follows: in Section \ref{sec:rel_works} related works on KGs for literary studies and on KG embeddings are surveyed. Section \ref{sec:semantic_model} describes the URW Ontology Network, on which the URW-KG relies. Section \ref{sec:kg} presents the data gathering and mapping pipeline. Section \ref{sec:kg_evl} is devoted to a quantitative evaluation of the resource based on KG embeddings. In Section \ref{sec:applications} potential applications of the KG are described. Future work and conclusion end the paper.

\section{Related Work} \label{sec:rel_works}

\subsection{Semantic Technologies for Literary Studies}

Several digital resources that provide information about literary works and writers are available online. Wikidata \cite{vrandevcic2014wikidata} 
is a general-purpose KG which includes knowledge about writers and their works. Other archives are domain-specific: Goodreads is a site owned by Amazon where readers share their impressions about books. Open Library is a project of the Internet  Archive\footnote{\url{https://archive.org/}}  where users can borrow books. Among these three archives, only Wikidata relies on the Linked Open Data paradigm. Open Library exposes its data through APIs, while Goodreads dismissed its APIs in $2020$. This leads to issues in data gathering and mapping, since there is not a unified model to align these resources.

Some digital archives are monographic and curated by teams of experts. It is the case of The European Literary Text Collection\footnote{https://www.distant-reading.net/eltec/} \cite{schoch2018distant}, a multi-lingual dataset of novels written from 1848 to 1920; DraCor\footnote{https://dracor.org/}  \cite{fischer2019programmable}, a collection of plays corpora in multiple languages; MiMoText\footnote{https://mimotext.github.io/}, a parallel corpus of French and German novels published from $1750$ to $1799$. 

Other resources are more oriented to explore the intersection between people and society. The Japanese Visual Media Graph\footnote{https://jvmg.iuk.hdm-stuttgart.de/} \cite{pfeffer2019japanese} gathers data about Japanese visual media (including manga and visual novels) from communities of fans. The Orlando Textbase\footnote{https://www.artsrn.ualberta.ca/orlando/} \cite{simpson2013xml} is a Knowledge Graph developed for exploring feminist literature. WeChangeEd\footnote{https://www.wechanged.ugent.be/} \cite{van2021women} is a Knowledge Graph of $1,800$ female editors born between 1710 and 1920, aligned with Wikidata.

The URW-KG is the first resource designed to study the intersection between literary production and ethnic information about writers. There are existing researches which analyzes the world of literature according to Wikipedia \cite{hube2017world}, but this is the first attempt to release a resource which could be at the same time a platform to foster DH and literary studies and a benchmark dataset for analyzing the nowledge gaps that affect an authoritative source like Wikidata in the literary domain.

\subsection{Knowlege Graph Embedding Techniques}
Knowledge Graph Embedding (GE) techniques aim at representing nodes and relations in a graph as dense vectors by projecting them into a vector space. In a nutshell, graph embedding techniques take as input the whole graph and generate as output a vector space representation for each node (and, sometimes, for each relation) in the graph. Applications of graph embedding range from predictions and clustering to entity linking and classification. 
Generally speaking, such techniques aim to preserve the original structure of the graph since nodes having similar neighborhood or nodes playing a similar role in the network have a similar representation \cite{Cai:18}. 

Several GE models have been proposed in the literature, and these can be grouped into 3 main categories \cite{sym13030485}: \textit{(i): Translation-Distance-Based Models}, which use a distance-based function for the scoring and a distance measure to evaluate predicted link in link prediction tasks; \textit{(ii): Semantic Information-Based Models}, which add some semantics information to the Translation-Distance-Based models, and so using more knowledge, to learn embeddings; \textit{(iii): Neural Network-Based Models}, which use neural networks to propagate neighborhood information among nodes and improve their representations.

In this work, we used four representative models for KG embeddings, \textit{i.e.,} TransE, TransR, DistMult and RESCAL.  TransE \cite{Bordes:13} is one of the most popular translation-based models which, given a triple in the form \textit{head - relation - tail} handles the relation as a translation from the head entity to the tail entity. As distance scoring function, given the triple \((h,r,t)\), it uses 
\( f(h,t) = |h+r-t| \).
Moreover, we also used TransR \cite{TransR:ref}. It is derived from TransE, but uses a specific relation-plane to handle different relations.
Next, we also considered DistMult \cite{distmult:ref}, a semantic information-based model whose key idea is to learn relation embedding with a bi-linear mapping function. In addition, semantic information derived from logical rules has been included in the embedding learning process, improving entities representations.
The last model used in this work is RESCAL \cite{rescal:ref}, another semantic information-based model; it uses a tensor factorization model that takes the inherent structure of relational data into account to perform the so-called \textit{collective learning}: this aims at adding information derived from attributes and relations of connected entities to support any kind of task (including node classification and link prediction).

In particular, we used these techniques to validate the design and the development of the URW-KG through both a qualitative and quantitative evaluation. More details will be provided next (Section \ref{ss:recommender}).
Up to our knowledge, the use of KG embedding techniques to investigate the underrepresentation of non-Western authors in recommendation and discovery tasks has never been investigated in literature.



\section{The UR Ontology Network} \label{sec:semantic_model}
The UR Ontology Network is a Semantic Model which includes a revised version of the Under-Represented Writers Ontology (URW-O) \cite{stranisci2021representing}\footnote{\url{purl.archive.org/urwriters}}, and a module dedicated to encoding writers' works, the Under-Represented Books Ontology (URB-O)\footnote{\url{purl.archive.org/urbooks}}. The former was specifically aimed at encoding biographical events and defining a robust criterion to identify writers suffering a potential lack of representation; the latter encodes knowledge about works and events related to them.
%
%
In the following, we first describe the criteria adopted for distinguishing writers on the basis of their underrepresentation. Then, a focus on intended usages and users is presented. The section concludes with a brief overview of the Ontology Network.

\subsection{Defining Underrepresentation} 
The purpose of the semantic model is to provide the adequate encoding for comparing Western and Transnational writers and their works with an emphasis on the potential lack of representation. The definition of underrepresentation itself is challenging since it may lead to arbitrary distinctions and to an inadequate terminology. In order to provide an objective and clear definition, we adopted the term `Transnational', which refers to people who ``operated outside their own nation’s boundaries, or negotiated with them'' \cite{boter2020unhinging}.  This definition focuses on the relation between a person and their country of birth, enabling the design of a taxonomy based on two axes: (i) the distinction between Western colonial and colonized countries, which follows the post-colonial theories \cite{spivak2015can}; (ii) the presence of ethnic minorities in a given Western country.
Transnational writers are defined as authors born in a former colony or belonging to an ethnic minority in a Western country and their works are classified consequently. Although such definition is not immune from false positives (e.g., Bernard-Henri Lévy, born in Algeri, or Italo Calvino, born in Cuba), the term `Transnational’ is adequately broad to justify their presence in such a category.  

\subsection{Intended Users and Usages}
The Semantic Model was designed bearing in mind two specific targets, listed in Table \ref{table:requirements}:  scholars in the humanities (usr$_1$) and computer scientists (usr$_2$). 
Despite not being the only potential users of the KG, both imply  represent specific instantiations of our
ROs.
%
The former may use the resource as a support for qualitative researches in world literature, post-colonial studies, and Digital Humanities; the latter may exploit the URW-KG as a benchmark for the development of book recommender systems, platforms for discovery and exploration of the authors and pipelines for the detection of biases in public sources of data. 

    
\begin{table}[ht]
\begin{center}
\begin{tabular}{l}
    \hline
          \textbf{Target End Users}\\
     \hline
     usr$_1$:   
     researchers in the humanities \\
     usr$_2$: computer scientists \\
     \hline
     \textbf{Intended Uses} \\
     \hline
     usg$_1$: supporting monographic world literature studies (usr$_1$) \\
     usg$_2$: tracking the social networks of writers (usr$_1$) \\
     usg$_3$: exploring biases in publicly available resources (usr$_2$)\\ 
     usg$_4$: building book recommender systems and exploration/discovery tools (usr$_2$) \\
    \vspace{0.05cm}
\end{tabular}
 \caption{A snapshot of the requirements for the semantic model}
 \label{table:requirements}
 \vspace{-1cm}
\end{center}
\end{table}

\subsection{Modeling Knowledge about Writers and Works}
\begin{figure}[t]
    \centering
    \includegraphics[width=0.8\textwidth]{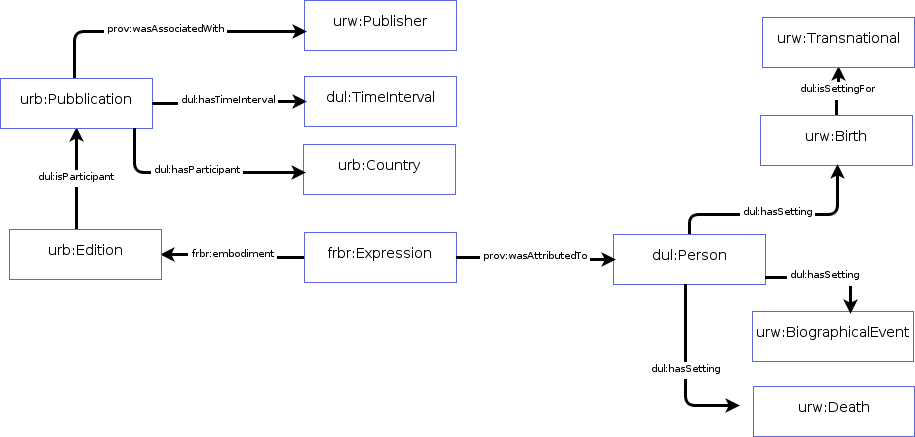}
    \caption{The basic patterns of the URW-O and URB-O ontologies. Prefixes are the following: ``dul'' stands dor DOLCE, ``frbr'' for the FRBR Ontology, ``prov'' for the PROV Ontology, ``urw'' for the URW-O and ``urb'' for the URB-O.}
    \label{fig:ontology_model}
    \vspace{-1mm}
    
\end{figure}

The UR Ontology Network consists of two distinct, but interconnected, ontologies: URW-O and URB-O. The former is aimed at representing biographical knowledge about writers and mainly relies on the TimeIndexedPersonStatus Ontology Design Pattern derived from DOLCE \cite{stranisci2021mapping}. As it can be observed in the right part of Figure \ref{fig:ontology_model}, biographical information about an author is not directly attached to them, but grouped into a \textsc{dul:Situation}, which is the setting for the author, their role and the other relevant information (eg: dul:TimeInterval, dul:Place).  \textsc{urw:Birth} is a special type of situation designed to distinguish between Western and Transnational writers. It is within this situation that such information is conveyed in the form of a role that a person has in relation with their birthplace. This modelling choice emphasizes the non-rigidity of such a condition: an author who is considered Transnational at their birth may change their status within their life.  \textsc{urw:Birth} is also mapped with the \textsc{crm:Birth} class formalized in the Cidoc-CRM ontology \cite{doerr2005cidoc}, a semantic model designed for providing information integration in the field of cultural heritage, and its biographical derivation, Bio CRM \cite{tuominen2018bio}. 

The URB-O inherits the encoding of works and editions from FRBR \cite{tillett2005frbr} and FaBiO \cite{peroni2012fabio}, designed for  guiding the cataloguing of written works according to Semantic Web principles (left part of Figure \ref{fig:ontology_model}). Since both ontologies consider the term `work' as an abstract concept rather than a tangible object, a work in our KG is modeled as an \textsc{frbr:Expression} which is \textbf{frbr:embodied} in one or more \textsc{urb:Edition}, namely a sub-class of the \textsc{frbr:Manifestation} class. 
Similarly to biographical events, we did not directly  attach publishing information to a book edition using properties, but defined a \textsc{urb:Publication} class where publication year, country, and publisher are grouped together with the edition. Technically, \textsc{urb:Publication} is a sub-class of \textsc{dul:Action}, from which the \textbf{dul:hasParticipant/dul:isParticipant} properties are inherited, and of  \textsc{prov:Activity}, which allows associating a publication with a publisher, through the property \textbf{prov:wasAssociatedWith}. The PROV Ontology \cite{lebo2013prov}, which provides a high-level set of classes and properties to specify the provenance of a given piece of information in a KG, is used to model the attribution of a book to their author (\textbf{prov:wasAttributedTo}), and to make explicit the source of knowledge from which an information is sourced (\textbf{prov:wasDerivedFrom}).

\section{Building the URW-KG} \label{sec:kg}

\subsection{Data Gathering and Analysis of UnderRepresentation}

To start, we gathered from Wikidata all the $393,441$ entities of type Person (wd:Q5) with occupation (wdt:P106)  writer (wd:Q36180), novelist (wd:Q6625963), or poet (wd:Q49757), with  their year of birth. We then filtered out all the people born before $1808$, which is a crucial year because it marks the beginning of the Spanish-American, which can be considered as the first decolonization process. The number of writers in our collection was thus reduced to $194,346$. Finally, for each author, we collected the country of birth, ethnic group, gender, date of death, Wikipedia page, and all the works associated with them. 
The country of birth has been used as a criterion for categorizing authors as Western or Transnational. Among the total number of writers on Wikidata, $176,697$ ($91\%$) of them are labeled as Western, while $17,368$ ($9\%$) are labeled as Transnational. The distribution is even more skewed when works are considered. Of $145,375$ works gathered from Wikidata, $136,995$ ($94.2\%$) belong to Western writers, whereas $8,380$ ($5.8\%$) are associated with Transnational Writers. 

\begin{figure}
    \includegraphics[width=\textwidth]{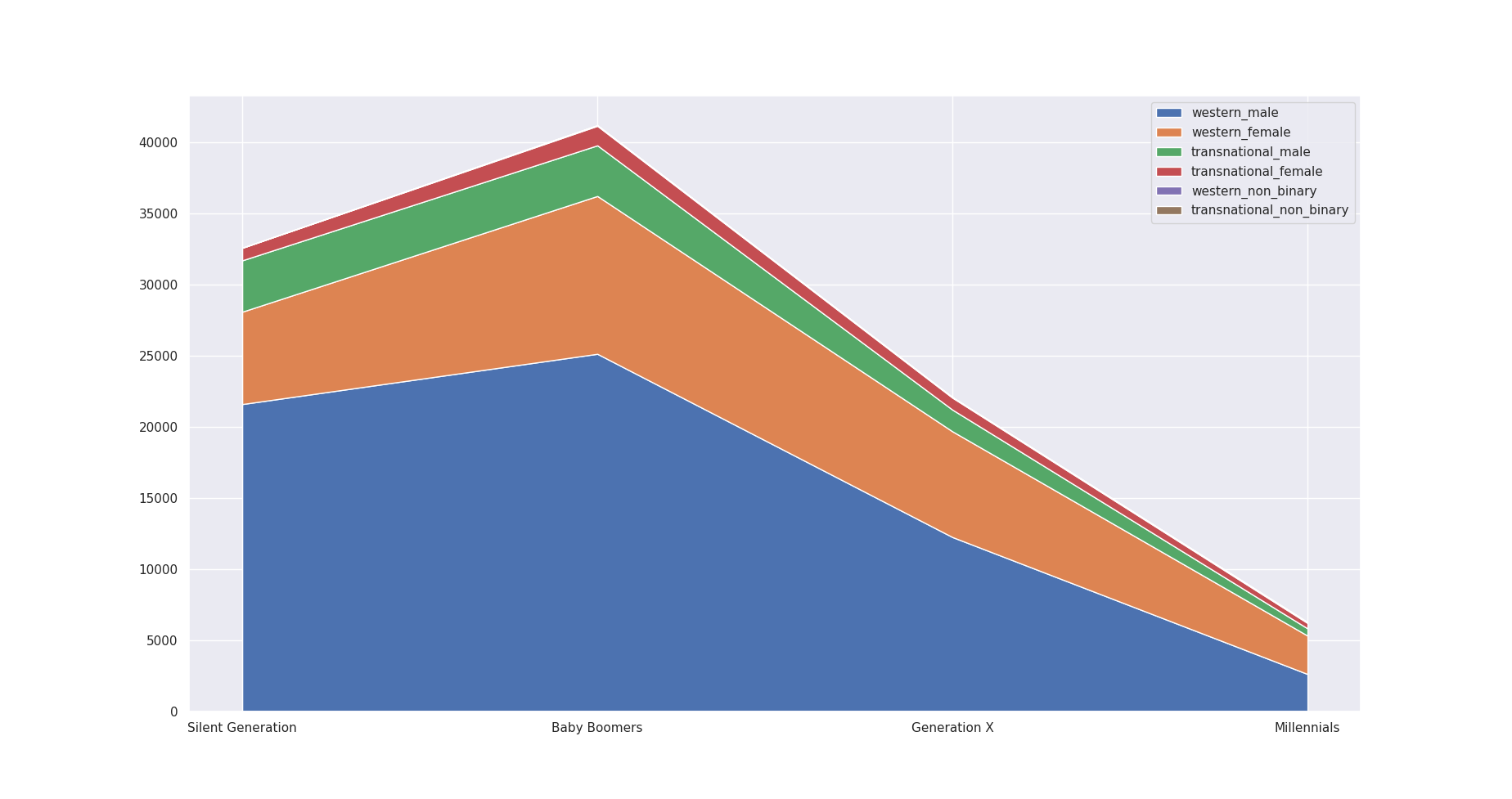}
    \caption{Four generations of Western and Transnational writers grouped by gender}
    \label{fig:generation}
\end{figure}

For better exploring such underrepresentation, we analyzed the distribution of Transnational and Western writers grouped by gender across four generations: Silent Generation (1928-1945), Baby Boomers (1946-1964), Generation X (1965-1980), and Millennials (1981-1996). As it can be observed in Figure \ref{fig:generation} Western male writers are predominant within the Silent Generation ($66.2\%$) and Baby Boomers ($60.9\%$). Surprisingly, the number of Western women progressively increases  until overcoming the number of men within the Millennials: $2,699$ ($43.1\%$) vs. $2,605$ ($41.6\%$). Transnational writers are significantly less across all the generations, but Transnational women writers suffer an additional lack of representation on Wikidata: there are only $508$ ($8.1\%$) male and $379$ ($6\%$) female Transnational writers on Wikidata among Millennials. Non-binary writers are the most underrepresented, regardless of their condition. In the KG there are only $146$ ($0.009\%$) non-binary Western authors and $23$ ($0.01\%$) non-binary Transnational authors.

\subsection{Mapping Wikidata with Other Sources of Knowledge}
A first strategy we adopted for mitigating the underrepresentation of Transnational writers on Wikidata was the enrichment of the KG through the integration with three external sources of knowledge: OpenLibrary\footnote{\url{https://openlibrary.org/}}, Goodreads\footnote{\url{https://www.goodreads.com/}}, and Google Books\footnote{\url{https://books.google.com/}}. The enrichment focused on works and not on authors, for two reasons: external sources are more prone to errors than Wikidata; considering writers as a constant allows evaluating the mitigation strategy in terms of the difference between the number of works in Wikidata and the number of works in external resources specialized in archiving books.

\paragraph{Mapping writers with external resources.} The first part of the mapping was devoted to collecting writers \textit{ids} from OpenLibrary and Goodreads. Although some Wikidata pages contain this information,  most of them do not. Hence, we adopted two heuristics for retrieving such information:
\begin{itemize}
    \item We retrieved all the names of the writers through the OpenLibrary APIs and kept only the entities fulfilling two conditions: (a) an exact string match between the author name in our KG and the one in OpenLibrary; (b) the same year of birth in our KG and in OpenLibrary. As a result, we obtained $55,834$ ids.
    \item We scraped all writers' names from Goodreads sitemap\footnote{\url{https://www.goodreads.com/siteindex.author.xml}} filtering out all homonyms. We then mapped all the names in our KG onto Goodreads author list, keeping only the string matches. We thus obtained $35,016$ ids.
\end{itemize}

\begin{figure}
    \centering
    \includegraphics[width=0.8\textwidth]{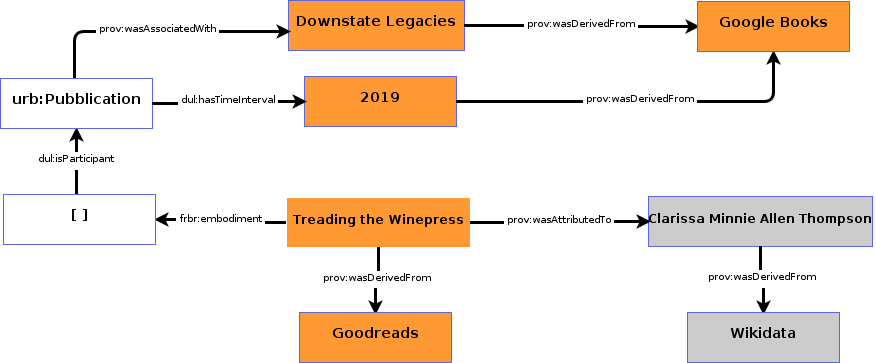}
    \caption{An enrichment example: works of Clarissa Minne Thompson Allen, missing from Wikidata, have been integrated into the URW-KG from Goodreads and Google Books.}
    \label{fig:ex_KG}
\end{figure}

\paragraph{Collecting works from Goodreads, OpenLibrary, and Google Books.} For the collection stage, we first considered all $55,834$ authors with an OpenLibrary id. Then, we retrieved from Goodreads all the $9,963$ writers who do not have an OpenLibrary id, instead of considering all authors with a Goodreads identifier. Such a choice was driven by the absence of APIs from Goodreads, which limits the information that can be collected from the platform and leads to a time consuming scraping process.
Given an author id, OpenLibrary APIs allow retrieving all works, and for each work it is possible to obtain all editions. Results include a set of useful information, among which: language, publisher, synopsis, subjects, year of publishing, place of publishing, ISBN. As a final result, $891,037$ works and $1,742,252$ editions were collected from OpenLibrary.
Goodreads does not provide APIs, but allows for web scraping. Hence, we first collected the list of all works from writers pages. Then, for each page we extracted the following information: year of publishing, synopsis, subjects, language, and ISBN. We also collected the  information about prizes and works, awarded and nominated. In total, $190,676$ works were collected from Goodreads.
After retrieving all works and editions from OpenLibrary and GoodReads, we searched additional information about them through the Google Books API, which was fed with all the collected ISBN. By doing so,  $1,177,801$ new items were obtained and mapped with OpenLibrary and Goodreads. 

Figure \ref{fig:ex_KG} shows the result of such a mapping on a specific writer: Clarissa Minnie Allen Thompsom, whose novel, Treading the Winepress; or, A Mountain of Misfortune,  was one of the first written by an African American woman. As it can be observed, Wikidata provides only information about her, but with the integration of knowledge from Goodreads and Google Books it is possible to obtain information about her work. 

\begin{figure}
    \centering
    \includegraphics[width=0.8\textwidth]{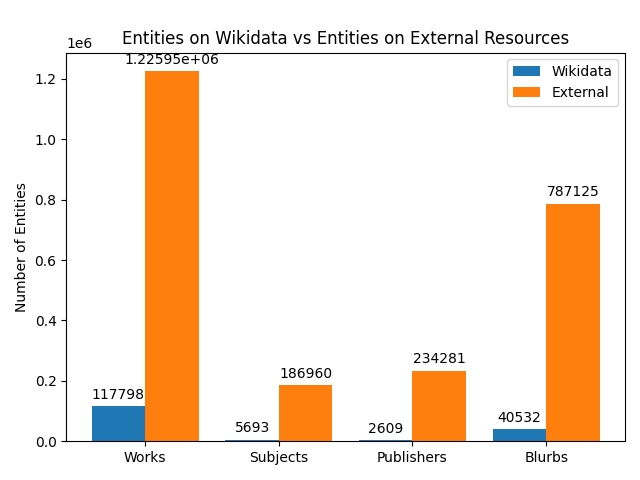}
    \caption{The impact of the mapping between Wikidata (in blue) and the other sources (Goodreads, Open Library, and Google Books, in orange) on $4$ types of entities: works, subjects, publishers, and blurbs.}
    \label{fig:bar_chart}
\end{figure}

\section{Evaluation of the Resource} \label{sec:kg_evl}
In this section, we first provide a quantitative overview of the quantity and characteristics of the works encoded in the KG, which show how the mapping resulted in a more balanced distribution between Western and Transnational works. Next, to further validate the principles that drove the creation of the resource, we present the output of some experiments based on KG embeddings techniques.\footnote{URW-KG and KG embeddings files: \url{https://doi.org/10.5281/zenodo.7450924}; \\Github folder: \url{https://github.com/marcostranisci/UnderRepresentedWritersProject}.}
The goal of these experiments is to show that URW-KG increases the probability for scholars and general audience to discover and be exposed to under-represented author and resources. 

\subsection{Impact of Data Integration on Underrepresentation}
A quantitative overview of the  information retrieved from external sources (Table \ref{table:augmented_data}, Fig. \ref{fig:bar_chart}) shows a significant increase of works (they are $16$ times more than in Wikidata) as well as an increase of the information about them. External resources include $787,125$ blurbs against the $40,532$ present in Wikipedia, and both the number of subjects and publishers extentively grow. 

\begin{table}[ht]
\centering
\begin{tabular}{l|r|r|r}
    \textbf{Entity type} & \textbf{Wikidata} & \textbf{External} & \textbf{Total} \\
    \hline
    Work & $117,798$ & $1,225,951$ & $2,050,726$ \\
    Edition & $0$ & $2,920,053$ & $2,920,053$\\
    Subject& $5,693$ & $186,960$ & $192,653$ \\
    Publisher& $2,609$ & $234,281$ & $236,890$\\
    Blurb & $40,532$ & $787,125$ & $827,657$ \\
\vspace{0.05cm}
\end{tabular}
 \caption{A comparison between the number of works, editions, subjects, publishers, and blurbs on Wikidata  and on external resources}
 \label{table:augmented_data}
\end{table}

\begin{figure}[h]
    \centering
    \includegraphics[width=0.7\textwidth]{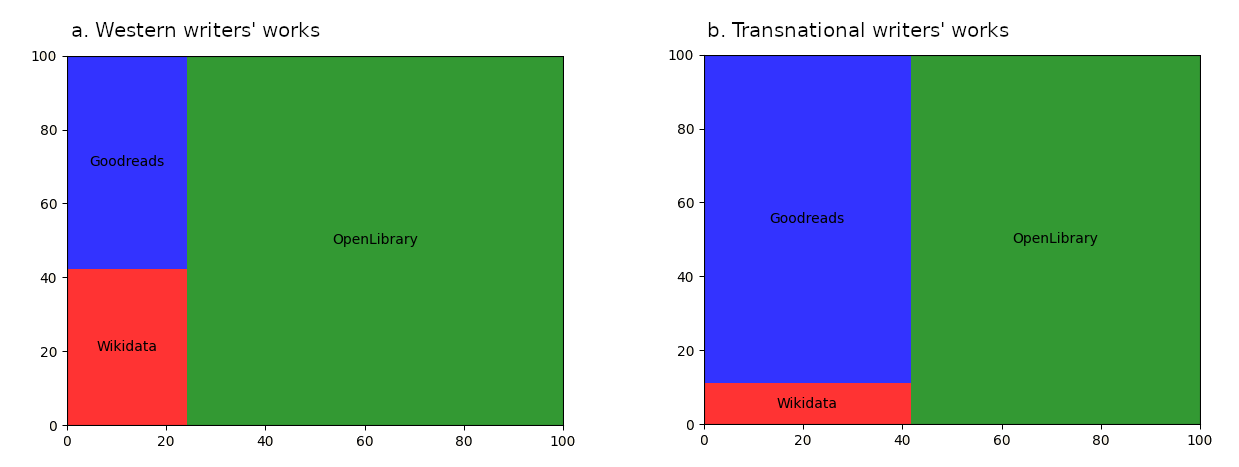}
    \caption{A comparison between the distributions of Western and Transnational works among Wikidata, Goodreads, and Open Library.}
    \label{fig:treemap}
\end{figure}

The increase of works after the integration of the additional sources not only has a general impact on the KG, but seems to mitigate the lack of representation of Transnational writers. The two treemaps in Figure \ref{fig:treemap} show that the impact of data from OpenLibrary and Goodreads is more significant for Transnational writers (b) than for Western (the red rectangle representing the percentage of Wikidata works is significantly smaller in (b) than in (a)). This means that the number of Transnational works gathered from external resources is higher, reflecting the wider preferences of readers and publishers in these crowdsourcing platforms. In fact, Transnational works grow from $8,380$ to $112,110$, while Western works increase from  $136,995$ to $1,113,841$,  reducing the Transnational:Western ratio from $1$:$16.3$ to $1$:$9.9$

\subsection{KG Evaluation through KG Embeddings} 
To further validate the effectiveness of the mapping between Wikidata and the other sources for the goal of mitigating the underrepresentation of non-Western writers and works, we carried out a series of experiments that exploit KG embedding techniques.
In particular, our experiments aimed at assessing to what extent the integration of new resources (\textit{i.e.,} OpenLibrary and GoodReads) allows users to be exposed to Transnational authors and works. Accordingly, we created three different portions of our KG, each of which is mapped to one of the original resources that compose URW-KG.


\begin{itemize}
    \item WD, which includes $24,224$ writers gathered from Wikidata (the $8.59\%$ are Transnational).
    \item OL, a sample of the KG derived from Open Library\footnote{Due to the computational load required to run graph embedding techniques on the complete portion, we selected a sample of authors that is representative of the resources encoded in the complete portion. In future work, we will replicate the experiments with the complete dataset.} which includes $25,049$ authors, of which $9.78\%$ are Transnational.
    \item GR, a portion of the KG with information gathered from Goodreads which  includes $9,838$ writers, of which $35.1$ are Transnational.
\end{itemize}

Next, we learned the embeddings by exploiting the techniques we previously mentioned (\textit{i.e.,} DistMult \cite{distmult:ref}, RESCAL \cite{rescal:ref}, TransE \cite{Bordes:13}, TransR \cite{TransR:ref}) on the three different portions of the graph. 
All the embeddings were generated by exploiting PyKeen \cite{pykeen:ref} Python library\footnote{\url{https://github.com/pykeen/pykeen}}. Dimension of the embeddings was set to 64, while we used default values for the other parameters.

After generating the embeddings, we selected a sample of $250$ Western writers from each dataset. Then, for each writer, we generated a list of similar authors by calculating the cosine similarity between the target writer and all the other authors available in the portion of KG. Finally, authors were ranked based on their descending similarity scores and the \textit{top-k} were selected for our analyses. 
The process was repeated on varying of the different portions of the graph and on varying of the different graph embedding techniques.

Finally, to evaluate to what extent the integration of new resources in URW-KG lead users to be more exposed to Transnational authors, we computed the ratio of Transnational writers who were among the top $1\%$, $5\%$, and $10\%$ of the lists of similar authors. Results are presented in Table \ref{table:embeddings_results}.
As regards graph embedding techniques, the differences among the approaches are tiny and probably not significant, but some interesting trends emerged anyway. TransE  was the best-performing technique on WD; however, when the portion of KG is switched to OL and GR, DistMult and RESCAL obtain the best performances. Accordingly, it is likely that both these techniques are the most effective at modeling and exploiting the information about Transnational works and authors.
By considering the differences on varying of the different portions of the KG, the experiments confirmed our intuitions. Indeed, when OL and GR are used as KG, the amount of Transnational authors put in the \textit{top-k} lists of similar authors is higher. This is true by considering both the percentage of Transnational authors over the total, as well as the rough amount of Transnational authors in the lists (put in parenthesis in Table \ref{table:embeddings_results}). 

Due to space reasons, we cannot provide further details about the behavior of the different graph embedding techniques. However, even these preliminary experiments provided us with encouraging findings, since they showed how the injection of new information in the KG and the selection of the most suitable graph embedding technique can be fundamental to expose users to new and unseen information about Transnational authors and works.
 

\begin{table}[ht]
\resizebox{1.0\textwidth}{!}{
\begin{tabular}{c|ccc|ccc|ccc}
    \multicolumn{1}{c|}{\textbf{Model}} & \multicolumn{3}{c|}{\textbf{Top 1\%}} & \multicolumn{3}{c|}{\textbf{Top 5\%}} & \multicolumn{3}{c}{\textbf{Top 10\% }} \\
    \hline
    
     & WD & OL & GR & WD & OL & GR & WD & OL & GR\\
     DistMult & 8.6\% (21) & \textbf{9.9\% (25)}& \textbf{34.8\% (34)} & 8.6\% (105) & \textbf{9.8\% (123)} & \textbf{35.4\% (174)} & \textbf{8.7\% (211)} & \textbf{9.8\% (247)} &  \textbf{35.2\% (347)} \\
     RESCAL & 8.3\% (20) & 9.6\% (24) & \textbf{34.8\% (34)} & 8.6\% (105) & \textbf{9.8\% (123)} & \textbf{35.4\% (172)} & 8.6\% (209) & 9.8\% (246) & \textbf{35.2\% (347)}\\
     TransE & \textbf{8.7\% (21)} & 9.3\% (23) & 32.6\% (32) & \textbf{8.7\% (106)} & 9.5\% (120) & 33.5\% (165) & \textbf{8.7\% (211)} & 9.6\% (241) & 33.6\% (330)\\
     TransR & 8.2\% (20) & 9.7\% (24) & 34.0\% (33) & 8.3\% (101) & 9.8 (123) & 34.7\% (171) & 8.3\% (203) & 9.8\% (246) & 35.0\% (344)\\
\vspace{0.05cm}    

\end{tabular}
}
\caption{KG embeddings results. }
 \label{table:embeddings_results}
\end{table}

\begin{figure}
    \centering
    \includegraphics[width=\textwidth]{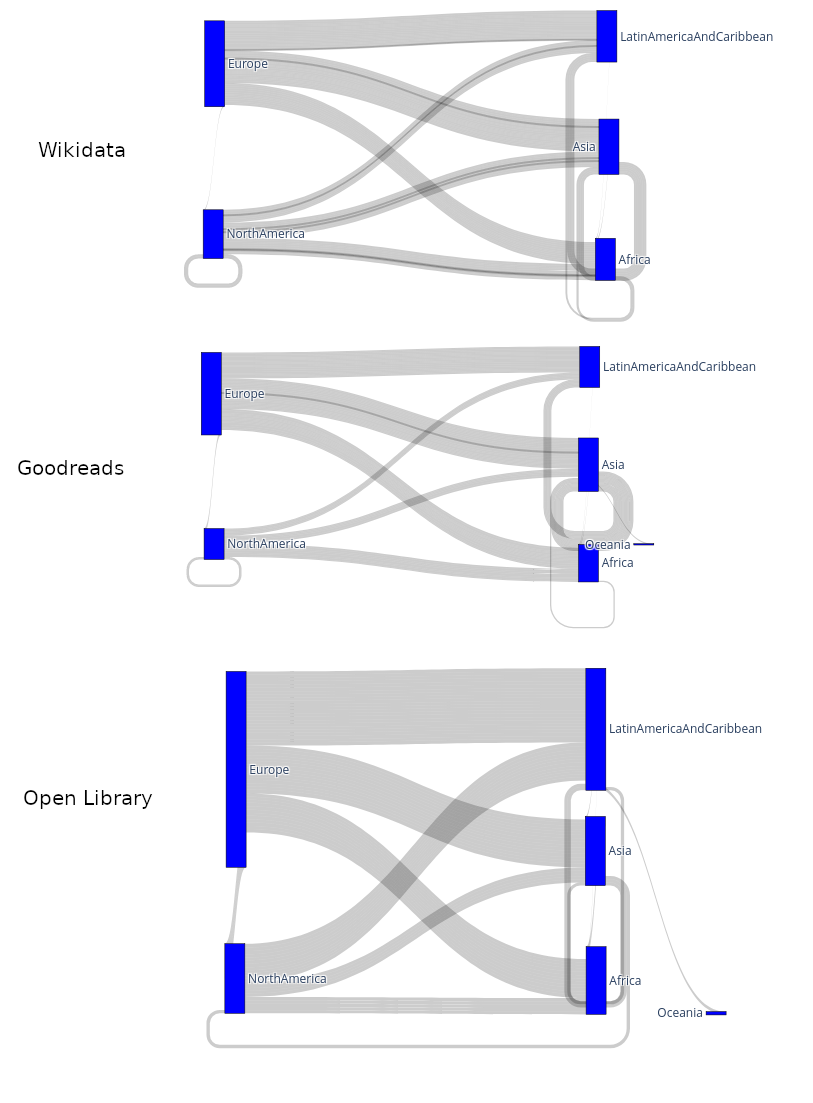}
    \caption{The relation between the continent of origin of Western and Transnational writers on Wikipedia, Goodreads, and Open Library.}
    \label{fig:links}
\end{figure}

Finally, KG embeddings were also exploited for investigating the interactions between authors' origins. Using the embeddings generated with DistMult, for each Western Writer of the previously selected samples we identified the Transational writer which is most similar to them, and retrieved the continent of origin of both authors. We thus obtained $250$ tuples of continents for each KG sample, that have been plotted in Figure \ref{fig:links}. As it can be observed, the geographical distribution of authors in these platforms is different. E.g. on Open Library there is an important stream between European and Latin American writers, while on Goodreads there are more connections between Asian and African writers. Such results show that different snapshots of the world literature may be derived from the three archives, and such differences seem to depend on the communities of users who generate contents on these platforms.

\section{Applications} \label{sec:applications}

In this section, an overview of possible applications of URW-KG is presented. In particular, we identified two main applications for our resource: recommender systems, and discovery tools for DH.

\subsection{Recommender Systems} \label{ss:recommender}

A straightforward application of URW-KG is represented by recommender systems \cite{resnick1997recommender}. Generally speaking, the goal of recommendation algorithms is to provide users with suggestions about new and unseen items they could be interested in. 
The effectiveness of this technology has been acknowledged for many years \cite{musto2022semantics}, and the domain of book recommendation is no exception. Indeed, several shreds of evidence \cite{Guo:20} showed that recommender systems based on KGs, such as URW-KG, can be helpful to provide people with accurate recommendations, also in the book domain.
However, most approaches currently presented focus on and exploit general-purpose KGs (such as KGs derived from or mapped to the Amazon catalogue). Differently from the current state of the art, the exploitation of URW-KG can lead to the development of new and different strategies to provide users with book recommendations, that are likely to expose them with new and unseen authors.
Indeed, one of the problems that currently affects recommender systems is the so-called popularity bias \cite{abdollahpouri2019popularity}, i.e., the tendency of recommendation algorithms to recommend very popular items to the majority of the users. Though relatively accurate, these recommendations are generally poorly surprising, and rarely expose users to new information. In this setting, the exploitation of resources such as URW-KG can be really helpful to mitigate this bias, since less trivial recommendations could be provided, thanks to the information encoded in the KG. As a side effect, the exploitation of our resource could also lead to an improvement in terms of fairness \cite{wang2022survey} of the recommendation lists, as well as in other \textit{beyond-accuracy metrics}\cite{Lops:19} such as the novelty, the diversity and the serendipity of the suggestions. As future work, a more detailed analysis of these aspects will be provided.


\subsection{Discovery tools for Digital Humanities}
The development of better systems for books and authors discovery may be of capital importance for  researcher in the humanities in several fields like post-colonial studies and world literature. 
%
After releasing the resource, we organized two structured interviews with a scholar in Contemporary Literature at the University of Birmingham and a scholar in Language and Translation studies specialized on African and Asian cultures at the University of Turin. 
The first focus of the interviews was the identification of a set of SPARQL queries that would fit their research needs. The following is a non-exhaustive list of suggestions emerged from the interviews:

\begin{enumerate}
    \item In how many language a work has been translated?
    \item How many editions of a work have been published through time and where?
    \item What has been published in a specific period and place?
    \item Is it possible to retrieve all the languages of the  works published in a given country?
    \item Is it possible to track a subject through time?
    \item Which are the social networks of authors or groups of authors?
\end{enumerate}

Most of these needs were immediately included as competency questions for querying the KG (suggestions $1$,$2$,$3$, and $5$). Suggestion $4$ could not be implemented because a distinction between  the first edition of a work and its translation is not yet possible due to the lack of such information on Open Library and Goodreads. The analysis of social networks ($6$) cannot be currently added to URW-KG, since the most knowledge about writers' life is stored in the form of raw texts, but can be addressed as described in \cite{DBLP:journals/corr/abs-2206-03547}.

The second focus of the interview was a qualitative assessment of KG embeddings results. We asked the scholars to lookup  a writer in the embeddings  and evaluate if the $10$ nearest authors were somehow correlated. Both found the results unexpected, although partly insightful, suggesting that more than a thematic or biographic similarity between authors, the KG embeddings provided a similarity based on the data source. 
Thereby, in order to make the KG embeddings useful for Humanities research activities, it is important to inject expert knowledge in the KG to identify academically relevant information, or to apply the KG embeddings techniques on curated portions of the KG.
%

\section{Conclusion and Future Work}
In this paper we presented the URW-KG, a dataset of writers and works built for tackling the underrepresentation of non-Western writers. The resource was developed by gathering and aligning data 
from Wikidata with three external resources: Goodreads, Google Books, and Open Library. Such an alignement has proven to be effective, as demonstrated through a quantitative overview of the graph and through a series of graph embedding experiments.

As future work, we will perform new KG embedding experiments on the whole dataset, with the goal of developing more inclusive and insightful discovery tools and recommender systems that rely on such resource.  
\bibliography{thebibliography}
\bibliographystyle{splncs04.bst}

\end{document}